\documentclass{article}

\usepackage{multirow}
\usepackage[final]{neurips_2022}

\usepackage{wrapfig}

\usepackage[colorlinks=true,linkcolor=blue, citecolor=blue]{hyperref}%
\usepackage[utf8]{inputenc} 
\usepackage[T1]{fontenc}    

\usepackage{url}            
\usepackage{booktabs}       
\usepackage{amsfonts}       
\usepackage{nicefrac}       
\usepackage{microtype}      
\usepackage{xcolor}         

\usepackage{graphicx}
\usepackage{amsmath}
\usepackage{amsthm}

\usepackage{physics}

\newcommand{\RB}{\mathbb{R}}

\newtheorem{thm}{Theorem}[section]
\newtheorem{rmk}[thm]{Remark}

\title{How You Start Matters for Generalization}

\author{%
 Sameera Ramasinghe$^{\dagger}$ \thanks{Corresponding email: sameera.ramasinghe@adelaide.edu.au} \hspace{1em} Lachlan MacDonald$^{\dagger}$ \hspace{1em}  Moshiur Farazi$^{\ddag}$ \\ 
    \AND
    Hemanth Saratchandran$^{\dagger}$
 \hspace{1em} Simon Lucey$^{\dagger}$ \\
 $^{\dagger}$University of Adelaide, $^{\ddag}$Australian National University
}

\begin{document}

\maketitle

\vspace{-2em}
\begin{abstract}
Characterizing the remarkable generalization properties of over-parameterized neural networks remains an open problem. In this paper, we promote a shift of focus towards initialization rather than neural architecture or (stochastic) gradient descent to explain this implicit regularization. Through a Fourier lens, we derive a general result for the spectral bias of neural networks and show that the generalization of neural networks is heavily tied to their initialization. Further, we empirically solidify the developed theoretical insights using practical, deep networks. Finally, we make a case against the controversial \emph{flat-minima} conjecture and show that Fourier analysis grants a more reliable framework for understanding the generalization of neural networks. 
\end{abstract}

\vspace{-1em}

\section{Introduction}
\label{sec:intro}
\vspace{-1em}
Neural networks are often used in the over-parameterized regime, and therefore, have the capacity to converge to many global minima that achieve zero training error. However, converging to such solutions is a non-convex, high-dimensional problem, which is typically intractable to solve. Furthermore, each of these minima may have unique properties that can lead to varying generalization performance, making some solutions preferred than others. Surprisingly, however, it is widely established that when neural networks are trained with gradient-based optimization techniques, they not only converge towards a global minimum, but also are biased towards solutions that exhibit good generalization even without explicit regularization. This mysterious behavior is flagged as the ``implicit regularization'' of neural networks and remains an open research problem.


To understand implicit regularization, numerous studies have considered simplified settings with restrictive assumptions such as linear networks \citep{ntk, soudry2018implicit, wei2019regularization, yun2020unifying, gunasekar2018implicit, wu2019implicit}, shallow networks \citep{gunasekar2018characterizing, ji2019implicit, ali2020implicit}, wide networks \citep{jacot2018neural, mei2019mean, chizat2020implicit, oymak2019overparameterized}, vanishing initialization \citep{chizat2019lazy, gunasekar2017implicit, arora2019implicit}, or infinitesimal learning rates \citep{ma2018implicit, li2018algorithmic, ji2018gradient, moroshko2020implicit}. Despite different assumptions, most of these works primarily focus on the effect of optimization procedure over the other factors and, at a high level, conclude that gradient-based optimizations guides neural networks toward max-margin solutions for separable data or minimize a notion of weight-norm in regression. While the aforementioned studies yield powerful insights, they lead to rather disconnected views that do not take important factors such as initialization or practical architectures  into consideration. As a result, faithful extrapolation of these theoretical conclusions to practical settings becomes problematic.


Our work is an attempt to fill this void. To this end,  we show substantial evidence that initialization plays a decisive role in determining the generalization of a neural network. In particular, we demonstrate that even with gradient-based optimization and a deep architecture -- networks can converge to solutions with extremely poor generalization properties. We further demonstrate that this result depends on the Fourier spectrum at initialization. \textbf{It should be noted that our result is not a recapitulation of the well-known observation that bad initialization hampers the convergence of neural networks. Rather, we show that initializing networks such that they have higher energies for higher frequencies leads to solutions that achieve perfect training accuracy, yet succumb to inferior test accuracy}. We further reveal that this is a generic property that holds in both classification and regression settings across various architectures. 





The roots of our analysis extend to the ``spectral bias'' (also known as the frequency principle) of neural networks \citep{xu2019training, rahaman2019spectral}. Spectral bias is an interesting phenomenon that implies networks tend to learn low frequencies faster, and consequently, tend to fit training data with low frequency functions. However, still there are some critical concerns entailed with existing research. First, studying spectral bias has so far been restricted to neural architectures with traditional activations such as ReLU \citep{rahaman2019spectral, xu2019training}, sigmoid \citep{luo2019theory}, or tanh \citep{xu2018understanding}. Thus, it is intriguing to investigate whether the spectral bias is a universal property that holds across various neural architectures and activation functions. This is important, as it enables the machine learning community to use spectral bias as a tool for investigating implicit regularization more systematically. Second, the question remains whether the spectral bias of a network is a sufficient condition for good generalization. We posit that the answers to these questions can potentially broaden the understanding of implicit regularization of neural networks.

The central objective of this paper aims  to shed light on the above questions. To this end, we utilize recently popularized implicit neural networks \citep{tancik2020fourier, ramasinghe2021beyond} (also referred to as coordinate-based networks) as an initial test-bed. Implicit neural networks are architecturally modified  fully-connected networks -- using non traditional activations such as Gaussians/sinusoids or positional embedding layers -- that can learn high-frequency functions rapidly. In particular, we first demonstrate that implicit neural networks do \emph{not} always converge to smooth solutions, contradicting mainstream expectations. In resolving this controversial observation, first, we invoke a \emph{compact data manifold hypothesis} to show that spectral bias is both architecture- and loss-agnostic in a general sense. To our knowledge, this is the most general proof of spectral bias to date in literature.  With this in hand, we affirm that the poor generalization of implicit neural networks is linked to the presence of high frequencies at initialization. Similarly, we further show that the implicit regularization of neural networks requires an initial spectrum that is biased towards lower frequencies. We postulate that the remarkable generalization properties of modern neural architectures can be partly attributed to the employment of non-linearities (such as ReLU) that exhibit such spectra upon random initialization. Extending the above analysis, we depict that even ReLU networks, when initialized with a higher frequencies, fail to converge to minima with good generalization properties.

Finally, we investigate the ``flat minima conjecture" (that the generalisation capacity of a neural network is determined by the flatness of the minimum to which training has converged) for multiple architectures.  We find that the consistency of the conjecture with experiment is architecture-dependent, while the predictions made using a spectral bias approach are consistent across all examined architectures and problems. 

Our main contributions are listed below:

\begin{itemize}
    \item We offer, to our knowledge, the simplest and most general proof of the spectral bias phenomenon in literature.  Our result applies to all presently-used neural network architectures (along with a vast space of parameterised models that are not neural networks) in problems satisfying the compact data manifold hypothesis.

    \item We show that initialization plays a crucial role in governing the implicit regularization of neural networks. Our results advocate for a shift of focus towards initialization in understanding the generalization paradox, which currently revolves around the optimization procedure. 
    
    
    
    \item  We conduct experiments in both classification and regression settings. We  show that the developed insights are generic across different architectures,  non-linearities, and initialization schemes. Our experiments include practical, deep networks, in contrast to many existing related works.
    
    \item We present (empirical) counter-evidence against the flat minima conjecture and show that 1) SGD is not always biased towards flat minima and 2) flat minima do not always correlate with better generalization. 
\end{itemize}

\section{Related Works}

\paragraph{Implicit regularization} Mathematically characterizing  implicit regularization of neural networks is at the heart of understanding deep learning. This intriguing phenomenon received increasing attention from the machine learning community after the seminal work by \citet{zhang2016understanding}, in which they showed that deep models, despite having the capacity to fit even \emph{random} data, demonstrate remarkable generalization properties. Since then, an extensive body of works have tried to characterize implicit regularization through various lenses including training dynamics \citep{advani2020high, gidel2019implicit, lampinen2018analytic, goldt2019dynamics, arora2019implicit}, flat minima conjecture \citep{keskar2016large, jastrzkebski2017three, wu2018sgd, simsekli2019tail, mulayoff2020unique}, statistical properties of data \citep{brutzkus2020inductive}, architectural aspects such as skip connections \citet{he2020resnet, huang2020deep}, and matrix factorization \citet{gunasekar2017implicit, arora2019implicit, razin2021implicit}. At a high-level, these works show that deep models implicitly minimize a form of weight norms, regularize derivatives of the outputs, or analogously, maximize a notion of margin between output classes. However, the center of attention of (almost all) these works is the bias induced by the optimization (SGD) methods. In contrast, we show that while SGD induces a crucial bias, (less highlighted) initialization also plays an important role. Notably,  \citet{min2021explicit} recently  discussed the role of initialization in the convergence and implicit bias of neural networks. They showed that the rate of convergence of a neural network depends on the level of imbalance of the initialization. Their setting, however, only considered single-hidden-layer linear networks under the square loss. In contrast, we offer richer insights and consider a broader and more practical setting in our work by using deeper,  practical networks. 

\vspace{-1em}
\paragraph{Spectral bias} Neural networks tend to learn low frequencies faster. To the best of our knowledge, this peculiar behavior was first systematically demonstrated on ReLU networks by \citet{xu2019training} and \citet{rahaman2019spectral} in independent studies, and a subsequent theoretical work  showed that shallow networks with Tanh activations \citep{xu2018understanding} also admit the same bias. Several recent works have also attempted to characterize the spectral bias of neural networks in different training phases and under various (relatively restrictive) architectural assumptions \citet{luo2019theory, zhang2019explicitizing, luo2020exact}. Perhaps, the insights developed by \citet{zhang2019explicitizing} and \citet{luo2020exact} are more closely aligned with some of the conclusions of our work, in which they showed that shallow ReLU networks with infinite width converge to solutions by minimally changing the initial Fourier spectrum.
\vspace{-1em}

\paragraph{Implicit neural networks} Implicit neural networks are a class of fully connected networks that were recently popularized by the seminal work of \citet{mildenhall2020nerf}. Implicit neural networks either use non-traditional activation functions (Gaussian \citep{ramasinghe2021beyond} or Sinusoid \citep{sitzmann2020implicit}) or positional embedding layers \citep{tancik2020fourier, zheng2021rethinking}. The key difference between implicit neural networks and conventional fully connected networks is that the former can learn high frequency functions more effectively and, thus, can encode natural signals with higher fidelity. Owing to this unique ability, implicit neural networks have penetrated many tasks in computer vision such as texture generation \citep{henzler2020learning,  oechsle2019texture, henzler2020learning, xiang2021neutex}, shape representation \citep{chen2019learning, deng2020nasa, tiwari2021neural, genova2020local, basher2021lightsal, mu2021sdf, park2019deepsdf}, and novel view synthesis \citep{mildenhall2020nerf, niemeyer2020differentiable, saito2019pifu, sitzmann2019scene, yu2021pixelnerf, pumarola2021d, rebain2021derf, martin2021nerf, wang2021nerf, park2021nerfies}.   

\section{Generalization and Fourier spectrum of neural networks}

\paragraph{Generalization of neural networks} Consider a set of training data $\{ \vb{x}_i, \vb{y}_i \}_{i=1}^N$ sampled from a distribution $\mathbb{D}$. Given a new set  $\{\bar{\vb{x}}, \bar{\vb{y}}\} \sim \mathbb{D}$, where a neural network $f$ only observes $\{\bar{\vb{x}}\}$, the goal is to learn a function such that $f(\bar{\vb{x}}) \approx \bar{\vb{y}}$. Since $\mathbb{D}$ is unknown, the network tries to learn a function that minimizes an expected cost  $\mathbb{E}[\mathcal{L}(f(\vb{x}_i), \vb{y}_i)]$ over the training data, where $\mathcal{L}$ is a suitable loss function. After training, if the network acts as a good estimator $f:\bar{\vb{x}} \to \bar{\vb{y}}$, we say that $f$ generalizes well. In classification, usually, a variant of the cross-entropy loss is chosen as $\mathcal{L}$, and in regression, $\ell_1$ or $\ell_2$ loss is chosen. It should be noted that generalization is entirely a function of $\mathbb{D}$ and thus, cannot be measured without priors on $\mathbb{D}$. In image classification, for instance, a held-out set of validation/testing data is used to as prior on $\mathbb{D}$ to measure the generalization performance. In regression, due to the infinite sampling precision of both input and output spaces, the use of such held-out data becomes less meaningful. Thus, a more  practical method of measuring the generalization in a regression setting, at least in an engineering sense, is to measure the ``smoothness'' of interpolation between training data. That is, we say that a network generalizes well if its output is smooth while fitting the training data. 


\paragraph{Smooth interpolations and the Fourier spectrum} In machine learning and statistics, a ``smooth'' signal is typically considered a signal with bounded higher-order derivatives. This interpretation stems from the fact that, in practice, noise causes large derivatives and, thus, suppressing higher-order derivatives is equivalent to suppressing noise in a signal, leading to better generalization. A widely used approach to obtain a smooth output signal is regularizing the second-order derivatives. For instance, in spline regression, a weighted sum of second-order derivatives and the square loss is minimized to achieve better generalization \citep{reinsch1967smoothing, craven1978smoothing, kimeldorf1970correspondence}. Interestingly, \citet{heiss2019implicit}, showed that shallow ReLU networks, when initialized randomly, implicitly regularize the second-order derivatives of the output over a broad class of loss functional, leading to better generalization. Next we show that minimizing the second-order derivatives of a signal is equivalent to minimizing the power of higher frequencies of that particular signal. Consider an absolutely integrable function $g(x)$ and its Fourier transform $\hat{g}(x)$. Then,

\[
  g(x) - \int_{-\infty}^{\infty} \hat{g}(k)e^{2\pi jkx} dk
\]

\[
\bigg| \frac{d^2 g(x)}{dx^2} \bigg| = \lvert 4\pi^2 \int_{-\infty}^{\infty} k^2\hat{g}(k)e^{2\pi jkx}dk \rvert \leq |4\pi^2| \int_{-\infty}^{\infty} |k^2\hat{g}(k)|dk
\]

Therefore, suppressing the higher frequencies of the Fourier spectrum $\hat{g}(k)$ of a signal reduces the upperbound on the magnitude of the second-order derivatives of that particular signal.

\vspace{-1em}
\paragraph{Fourier spectrum of a neural network} To any integrable function $f:\mathbb{R}^d\rightarrow\mathbb{R}$ is associated its \emph{Fourier transform}, given by the formula $\mathcal{F}[f](\vb{k}):=\int e^{-i\vb{k}\cdot\vb{x}}f(\vb{x})\,d\vb{x}$ \cite{classicalfourier}.  In particular, a scalar-valued neural network defines a function $f_{\theta}:\mathbb{R}^d\rightarrow\mathbb{R}$, whose Fourier transform makes sense provided $f_{\theta}$ is integrable.  We will mollify (set to zero outside of some set) $f_{\theta}$ to take into account data locality, which guarantees integrability.  The Fourier transform of a vector-valued network is defined by taking the Fourier transform of each of its component functions.

\begin{figure}[!htp]
\centering
\includegraphics[width=1.\columnwidth]{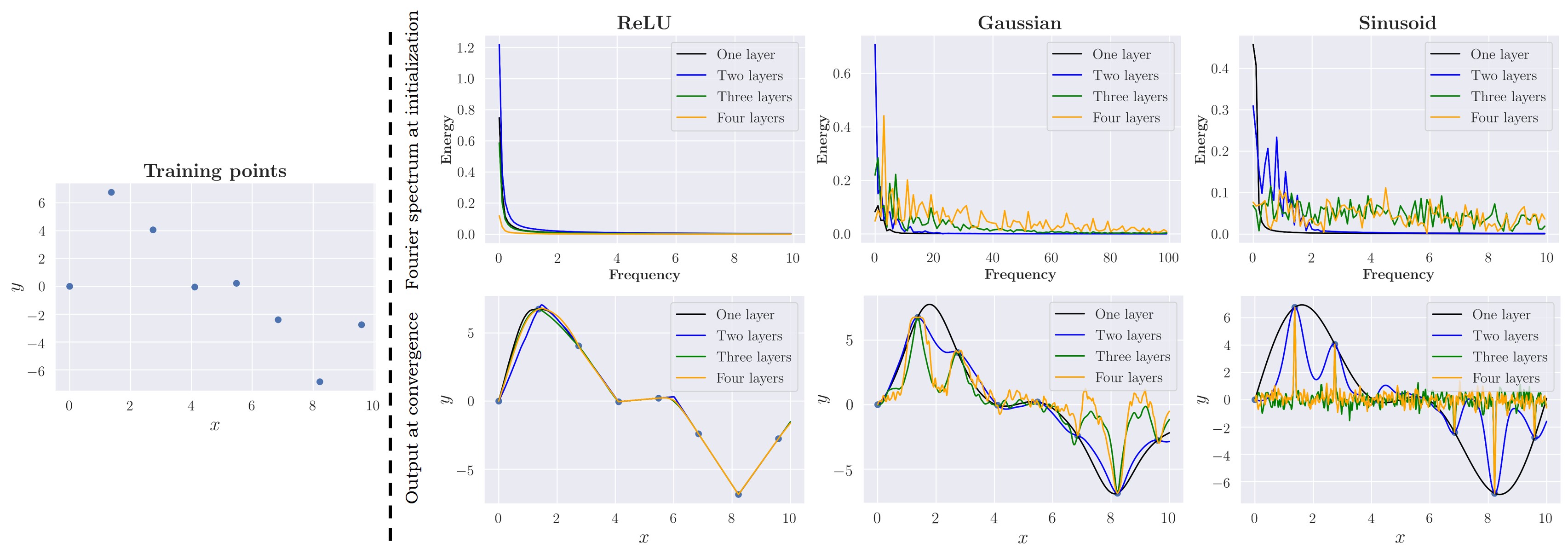}
\vspace{-2em}
\caption{\small \textbf{Implicit neural networks are not implicitly regularized.} The ReLU network keeps converging to smooth solutions despite the increasing depth. In contrast, Gaussian and sinusoidal networks converge to increasingly erratic solutions as the depth is increased. Interestingly, note that the  Gaussian and sinusoidal networks add higher frequencies to the spectrum at initialization as the depth is increased, in contrast to the ReLU network.}
\label{fig:depth_interp}
\end{figure}

\vspace{-1em}
\section{Implicit neural networks do not always generalize well}
\label{sec:implicit}

In this section, we compare implicit neural networks against conventional ReLU networks in regression, and show that the former do not always generalize well. The experiments are described in detail below.


\paragraph{Experiment 1:} We utilize fully-connected networks with three types of activation functions: 1) ReLU, 2) Gaussian, and 3) Sinusoidal. We sample $8$ sparse points from the signal $3\mathrm{sin}(0.4\pi x) + 5\mathrm{sin}(0.2\pi x)$ and regress them using networks across varying depths.  As depicted in Fig. \ref{fig:depth_interp}, when more capacity is added to the ReLU network via hidden layers, the network keeps converging to a smooth solution as expected. In contrast, Gaussian and Sinusoidal networks showcase worsening interpolations, contradicting the mainstream expectations of implicit regularization. Interestingly, it can be observed that sinusoidal and Gaussian networks add more energy to higher frequencies at initialization as more layers are added. In contrast, ReLU networks tend to have a highly biased spectrum towards lower frequencies irrespective of the depth. All the networks are randomly initialized using Xavier initialization \citep{glorot2010understanding}. We use  SGD to optimize the networks with a learning rate of $1 \times 10^{-4}$. The networks consist of $256$ neurons in each hidden layer.


Experiment 1 concludes that even with SGD as the optimization algorithm, not all types of networks are implicitly regularized. Instead, the results hint that the initial Fourier spectrum impacts the generalization performance of a neural network, and the network architecture (activation) plays a crucial role in determining the spectrum. In the upcoming sections, we dig deeper into these insights.


 

\section{The universality of spectral bias}
\label{sec:universality}
\vspace{-1em}
Sec.~\ref{sec:implicit} showed that  networks with higher frequencies at initialization tend to exhibit poor generalization. However, it is worth investigating if there is indeed a causal link between the two. Intuitively, spectral bias allows us to speculate such a link. That is, one can speculate that the non-smooth interpolations are a result of unwanted residual frequencies after the convergence of lower frequencies. Continuing this line of thought, we present a general proof of the spectral bias, and show that spectral bias is a universal phenomenon that exists in any parameterized function (which  includes the class of all neural networks), given that they are trained with  gradient-based optimization methods. 

Let $f:\RB^p\times \RB^d\rightarrow\RB$ be a parameterised family $\theta\mapsto f_{\theta}$ of continuous functions $\RB^d\rightarrow\RB$.  We assume that the map $(\theta,x)\mapsto f_{\theta}(x)$ is differentiable almost everywhere, and that the restriction of the (almost everywhere-defined) map $x\mapsto D_{\theta}f_{\theta}(x)$ is bounded over any compact set.  This setting includes all presently used neural network architectures, with activation functions constrained only to be differentiable almost everywhere.

We care only about the behaviour of $f_{\theta}$ in a neighbourhood of the data.  We invoke the \emph{compact data manifold hypothesis}: that the entire data manifold is contained in some compact neighbourhood\footnote{A \emph{compact neighbourhood} is a compact set containing a nonempty open set.} $K$. Let $g_{\theta}$ be the extension by zero of $f_{\theta}$ outside of $K$, i.e.
\begin{equation}
g_{\theta}(x):=\begin{cases}f_{\theta}(x) &\text{ if $x\in K$},\\ 0&\text{ otherwise}\end{cases}
\end{equation}
Thus $g_{\theta}$ has compact support\footnote{The \emph{support} of a function is the smallest closed set containing the set on which the function is nonzero} $K$ and is continuous on $K$ since $f_{\theta}$ is continuous globally.  It follows that $g_{\theta}$ is in $L^1(\RB^d)$.  It follows from the Riemann-Lebesgue lemma \cite[Proposition 2.2.17]{classicalfourier} that the Fourier transform $\mathcal{F}[g_{\theta}]$ of $g_{\theta}$ vanishes at infinity.  The next theorem shows that the same is true of the change $\frac{d}{dt}[g_{\theta(t)}]$ during training, yielding the first completely general proof of the spectral bias.

\begin{thm}[The spectral bias of  differentiably parameterised models]\label{fprinc}
Let $c:\RB\times\RB\rightarrow\RB$ be any differentiable cost function, and let $\{x_{i}\}_{i=1}^{n}$ be a training set drawn from the data manifold, with corresponding target values $\{y_{i}\}_{i=1}^{n}$.  Assume that the parameterised function $\theta\mapsto g_{\theta}$ is trained according to almost-everywhere-defined gradient flow:
\begin{equation}
\frac{d}{dt}g_{\theta(t)}(x) = -\frac{1}{n}\sum_{i=1}^{n}\mathcal{K}(\theta(t),x,x_{i})\nabla c(g_{\theta(t)}(x_{i}),y_{i}),
\end{equation}
where
\begin{equation}
\mathcal{K}(\theta,x,x'):=D_{\theta}g_{\theta}(x)\,D_{\theta}g_{\theta}(x')^T
\end{equation}
is the tangent kernel \cite{ntk} defined by $g_{\theta}$.  Then the Fourier transform $\mathcal{F}[g_{\theta(t)}]$ evolves according to the differential equation
\begin{equation}
\frac{d}{dt}\mathcal{F}[g_{\theta(t)}](\xi) = -\frac{1}{n}\sum_{i=1}^{n}\int_{x\in\RB^d}e^{-ix\cdot\xi}\mathcal{K}(\theta(t),x,x_{i})\nabla c(g_{\theta(t)}(x_{i}),y_{i})\,dx.
\end{equation}
Moreover, $\frac{d}{dt}\mathcal{F}[g_{\theta(t)}]$ vanishes at infinity: $\big|\frac{d}{dt}\mathcal{F}[g_{\theta(t)}](\xi)\big|\rightarrow 0$ as $\|\xi\|\rightarrow\infty$.
\end{thm}

Two remarks are in order regarding the antecedents of Theorem \ref{fprinc}.

\begin{rmk}\normalfont
Our use of the tangent kernel in characterising the dynamics of gradient flow are inspired by the seminal work of \cite{ntk}, which is well-known for hypothesising infinite width for several of its results.  In fact, the tangent kernel governs gradient flow dynamics \emph{independently of any architectural assumptions} (beyond the stated differentiability assumption), and in particular, Theorem \ref{fprinc} does not require an assumption of infinite width in order to use the tangent kernel.  The infinite width hypothesis is invoked in \cite{ntk} specifically to give a simple proof of evolution towards a global minimum.  We do not attempt any such proof and so do not require the infinite width hypothesis.
\end{rmk}

\begin{rmk}\normalfont
Our second remark concerns previous proofs of the frequency principle \citep{zhang2019explicitizing, luo2020exact}, wherein it was deemed necessary to make simplifying assumptions such as limited depth and infinite width.  With these assumptions the authors are able to analytically compute decay rates for the spectral bias.  These works also do not take into account the structure of the data.  An alternative, more general approach using Sobolev space theory is taken in \cite{luo2019theory}, wherein the data distribution is accounted for.  While \cite{luo2019theory} also gives explicit decay rates, the proofs are relatively involved and are only given for chain network architectures.  In contrast, we abstract away all detail concerning model architecture beyond minimal regularity assumptions that hold for all presently used neural network models, thereby exhibiting the spectral bias to hold generically for parameterised models, greatly simplifying its proof.
\end{rmk}

\begin{figure}[h]
\includegraphics[width=1.0\columnwidth]{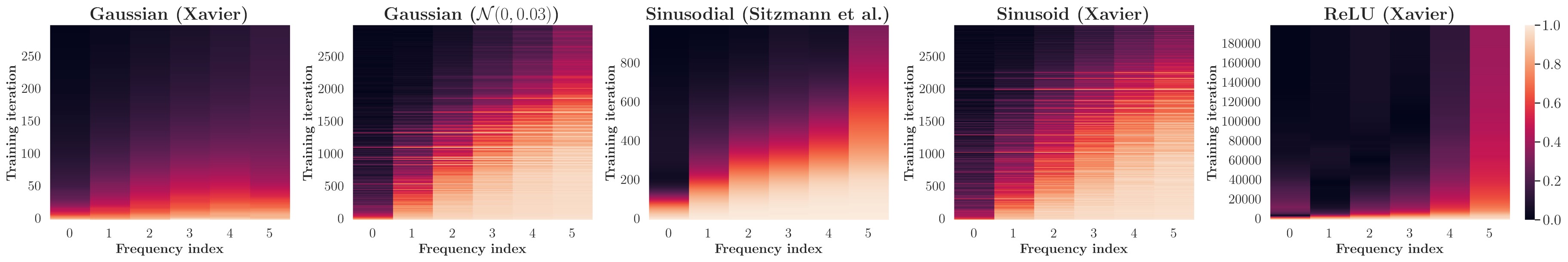}
\vspace{-2em}
\caption{\small \textbf{Spectral bias applies to different network types and initialization schemes.} We measure the convergence of each frequency index as the training progresses. The colors indicate the difference between the ground truth and the predicted frequencies at each index. Xavier and Sitzmann are the initialization schemes proposed by \citet{glorot2010understanding} and \citet{sitzmann2020implicit}, respectively. Note that the convergence-decay rates of frequencies varies across network types and initialization schemes.}
\label{fig:spectral_bias}
\end{figure}

\textbf{Experiement 2: } The goal of this experiment is to (partially) empirically validate the above theoretical conclusions. To this end, we use ReLU, Gaussian, and sinusoid  networks. We train the networks on densely sampled points from $g(x) = \sum_{n=1}^{6} \mathrm{sin}(10\pi nx )$.  While training, we visualize the convergence of frequency indices of all the networks (Fig. \ref{fig:spectral_bias}). As Theorem \ref{fprinc} predicted, all three types of networks exhibit spectral bias. Note that the convergence-decay rates differ across network-types and initialization schemes, which also has an impact on generalization (see Appendix).



In the next section, we show that the initialization plays a key role in generalization and the widely-observed good generalization properties of ReLU networks are merely a consequence of them having biased initial spectra (towards lower frequencies), upon random initialization. 

\vspace{-1em}
\section{ReLU networks do not always generalize well}
\vspace{-1em}
In this section, we show that the initial Fourier spectrum plays a decisive role in governing the implicit regularization of a neural network. Notably, we show that even ReLU networks (which are commonly expected to regularize well) do not always converge to smooth solutions despite training with SGD. 

\paragraph{Experiment 3a:} We investigate and analyze the effect of the initial Fourier spectrum on generalization. First, we sample a signal $\mathrm{sin}(\pi x)$ with a step size of $1$. Thus, the lowest frequency signal that can fit this set of training points is $\mathrm{sin}(\pi x)$ (Nyquist-Shannon sampling theorem). Then, we randomly initialize a ReLU network using Xavier initialization, so that its initial Fourier spectrum does not contain significant energies above the frequency index $k = 0.5$ (which corresponds to the lowest frequency solution). After training the network over the training points, the network converges to the lowest frequency solution, \textit{i.e.}, $\mathrm{sin}(\pi x)$. 

\paragraph{Experiment 3b:} We utilize the same training points used in Experiment 4a. However, in this instance, we pre-train the ReLU network on a signal $\mathrm{sin}(10 \pi x)$. Note that at this instance, the Fourier spectrum of the network has a spike at $k=5$, which is above $k = 0.5$. Then, starting from these pre-trained weights, we train the network on the training points. 

\paragraph{Experiment 3c:} We initialize a Gaussian network with Xavier initialization, so  that it contains frequencies above $k=0.5$. Then, starting from these  weights, we train the model on the above training points.

\paragraph{Experiment 3d:} We initialize a Gaussian network with a random weight distribution $\mathcal{N}(0,0.03)$ such  that it does not contain frequencies above $k=0.5$. Then, starting from these  weights, we train the model on the training points used in the above experiments.

\begin{figure}[h]
\centering
\includegraphics[width=1.\columnwidth]{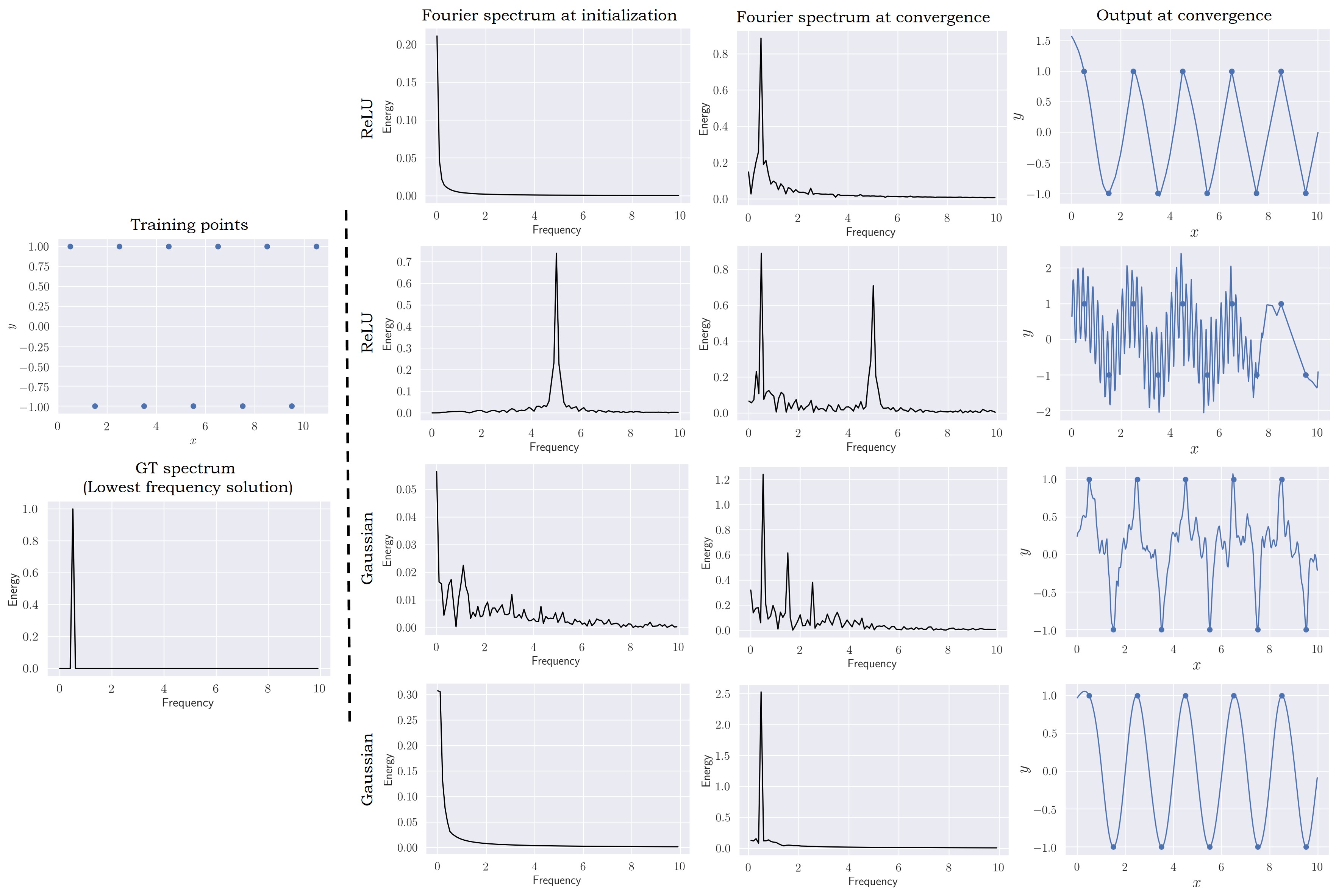}
\vspace{-1em}
\caption{\small Left block shows sparsely sampled training points from $\mathrm{sin}(\pi x)$ and the corresponding lowest frequency solution that fits the training data. Right block compares generalization corresponding to different networks and initializations.  \textit{Top row: } The ReLU network tries to converge to a solution by changing low frequencies at a faster rate due to spectral bias. Consequently, when initialized with no high frequencies, the network ends up converging to the lowest frequency (hence smooth) solution for the training points. \textit{Second row: } \textbf{ReLU networks do not always generalize well.}  If higher frequencies (than the lowest frequency solution) exist at initialization, ReLU networks reach a solution manipulating only the lower frequencies, resulting in bad interpolations. \textit{Third row: } Same behavior is demonstrated with a Gaussian network. \textit{Fourth row: } \textbf{Gaussian networks can generalize well if initialized properly.} Since the network does not contain high frequencies at initialization, it is possible for the network to converge to the lowest frequency solution.}
\label{fig:initialization}
\end{figure}

Fig.~\ref{fig:initialization} visualizes the results. As illustrated, when the spectrum of the ReLU network does not contain frequencies higher than $k=0.5$, the final spectrum of the network matches with the lowest frequency solution. In contrast, when the initial spectrum of the ReLU network contains frequencies higher than $k=0.5$, the network adds a spike at $k=0.5$, but leaves the high-frequency spike untouched as the network has already reached zero train error. This results in a non-smooth (poorly generalized) solution. Interestingly, observe that Gaussian networks also can generalize well if the initial spectrum does not contain higher frequencies. It is vital to note that, however, the convergence-decay rates of frequencies also play an important role. For instance, if the convergence-decay rate is low, higher frequencies begin to get affected \emph{before} the lower frequencies are converged, which can lead to non-smooth solutions (see Appendix).  In the next section, we investigate the effect of having high bandwidth spectra at initialization in classification, using popular deep networks.

\vspace{-1em}
\section{Generalization of deep networks in classification}
\vspace{-1em}
Sec. \ref{sec:universality} affirmed that the spectral bias holds for \emph{any} parameterized model trained using gradient descent.  Thus, it is intriguing to explore whether the practical insights we developed thus far extrapolate to popular deep networks that are ubiquitously used. However, for deep networks with high-dimensional inputs (\textit{e.g.}, images), high-bandwidth initialization becomes less straightforward. For instance, consider a network that consumes high dimensional inputs $\vb{x} = (x_1, x_2, \dots, x_n)$. Then, one can hope to directly extend the one dimensional technique we used and train the network on the supervisory signal $\mathrm{sin}(wx_1) \times \mathrm{sin}(wx_2) \times \dots \times \mathrm{sin}(wx_n)$. Nevertheless, it is easy to show that in this case, as the dimension of the input grows, the target signal converges to zero. Therefore, we opt for an alternative method and pre-train the models on random labels to obtain higher bandwidths (see Appendix for a discussion).

\paragraph{Experiment 4} We use $9$ popular models for this experiment: VGG16, VGG11, AlexNet, EfficientNet, DenseNet, ResNet-50, ResNet-18, SENet, and ConvMixer. In the first setting, we initialize the models with random weights, train them on the train splits of the datasets, and measure the test accuracy on the test splits.   In the next setting, we first pre-train the models on the train split with randomly shuffled labels. Then, starting from the pre-trained weights, we train the models on the train splits of datasets with correct labels and compute the test accuracy on the test splits. The results are depicted in Table \ref{tab:deep_networks}.

Recall that the pre-trained models on random labels yield higher initial bandwidths compared to randomly initialized models. As evident from the results, starting from a higher bandwidth hinders good generalization, validating our previous conclusions. The test accuracies of some models under random weight initialization (Table \ref{tab:deep_networks}) are slightly lower than the benchmark results reported in the literature. This is because, following \citet{zhang2016understanding}, we treat data augmentation as an explicit regularization technique and do not use it. In contrast to \citet{zhang2016understanding}, however, we consider dropout and batch normalization as architectural aspects and keep them.

\begin{table}[!htp]
\scriptsize
\centering
       \begin{tabular}{||c|c|c|c|c||}
\hline

\multicolumn{5}{||c||}{CIFAR10} \\

\hline
\multirow{ 2}{*}{Model} & 
\multicolumn{2}{|c|}{Random initialization}  &  \multicolumn{2}{|c||}{High B/W initialization} \\
\cline{2-5}
 & 
 Train accuracy & Test accuracy & Train accuracy &  Test accuracy \\
\hline
VGG11 \citep{simonyan2014very}              & $100\%$ & $84.33 \pm 0.49\%$                   & $100\%$ & $71.94 \pm 0.71\%$ \\
VGG16 \citep{simonyan2014very}              & $100\%$ & $88.24 \pm 0.12\%$          & $100\%$ & $71.55 \pm 0.79$\%\\
AlexNet \citep{krizhevsky2014one}           & $100\%$ & $80.11 \pm 1.13\%$                   & $100\%$ & $51.31 \pm 0.61\%$ \\
EfficientNet \citep{tan2019efficientnet}    & $100\%$ & $76.78 \pm 0.57\%$          & $100\%$ & $61.38 \pm 0.46\%$\\
DenseNet \citep{huang2017densely}           & $100\%$ & $86.69 \pm 0.02\%$          & $100\%$ & $80.86 \pm 0.01\%$\\
ResNet-18 \cite{he2016deep}                  & $100\%$ & $82.44 \pm 0.15\%$                   & $100\%$ & $68.99 \pm 0.62\%$\\
ResNet-50 \cite{he2016deep}                  & $100\%$ & $87.18 \pm 0.21\%$          & $100\%$ & $62.72 \pm 0.47\%$ \\
SENet \citep{hu2018squeeze}                 & $100\%$ & $86.31 \pm 0.30\%$          & $100\%$ & $71.20 \pm 0.35\%$\\
ConvMixer \citep{trockman2022patches}       & $100\%$ & $86.72 \pm 0.97\%$                   & $100\%$ & $49.33 \pm 0.78\%$\\
\hline
\hline
\multicolumn{5}{||c||}{CIFAR100} \\

\hline
\multirow{ 2}{*}{Model} & 
\multicolumn{2}{|c|}{Random initialization}  &  \multicolumn{2}{|c||}{High B/W initialization} \\
\cline{2-5}
 & 
 Train accuracy & Test accuracy & Train accuracy &  Test accuracy \\
\hline
VGG11                   & $100\%$ & $54.03 \pm 0.71\%$                 & $100\%$ & $41.88 \pm 0.94\%$\\
VGG16                   & $100\%$ & $56.86 \pm 0.68\%$      & $100\%$ & $36.27 \pm 1.92\%$\\
AlexNet                 & $100\%$ & $53.12 \pm 1.01\%$                 & $100\%$ & $41.44 \pm 1.21\%$ \\
EfficientNet            & $100\%$ & $43.15 \pm 0.58\%$      & $100\%$ & $26.83 \pm 0.89\%$ \\
DenseNet.               & $100\%$ & $57.76 \pm 0.24\%$      & $100\%$ & $46.56 \pm 0.41\%$\\
ResNet-18                & $100\%$ & $52.14 \pm 0.61\%$                 & $100\%$ & $41.98 \pm 0.59\%$  \\
ResNet-50                & $100\%$ & $54.42 \pm 0.78\%$      & $100\%$ & $30.56 \pm 0.58\%$  \\
SENet                   & $100\%$ & $58.64 \pm 0.25\%$      & $100\%$ & $51.56 \pm 0.77\%$\\
ConvMixer               & $100\%$ & $61.20 \pm 0.24\%$                 & $100\%$ & $26.35 \pm 0.99\%$\\
\hline
\hline
\multicolumn{5}{||c||}{Tiny ImageNet} \\

\hline
\multirow{ 2}{*}{Model} & 
\multicolumn{2}{|c|}{Random initialization}  &  \multicolumn{2}{|c||}{High B/W initialization} \\
\cline{2-5}
 & 
 Train accuracy & Test accuracy & Train accuracy &  Test accuracy \\
\hline
VGG11               & $100\%$ & $38.87 \pm 0.41\%$               & $100\%$ & $27.99 \pm 0.73\%$   \\       
VGG16               & $100\%$ & $40.95 \pm 0.61\% $               & $100\%$ & $21.77 \pm 0.85\%$\\
AlexNet             & $100\%$ & $35.56 \pm 0.55\%$               & $100\%$ & $21.94 \pm 1.66\%$   \\
EfficientNet        & $100\%$ & $33.39 \pm 0.42\%$      & $100\%$ & $18.19 \pm 0.92\%$   \\
DenseNet            & $100\%$ & $48.86 \pm 0.35\%$      & $100\%$ & $28.23 \pm 0.27\%$   \\
ResNet-18            & $100\%$ & $43.58 \pm 1.21\%$               & $100\%$ & $26.73 \pm 0.63\%$ \\
ResNet-50            & $100\%$ & $43.33 \pm 1.43\%$      & $100\%$ & $28.08 \pm 0.61\%$   \\
SENet               & $100\%$ & $28.27 \pm 1.33\%$      & $100\%$ & $24.03 \pm 0.29\%$   \\
ConvMixer           & $100\%$ & $45.38 \pm 0.88\%$               & $100\%$ & $27.77 \pm 0.28\%$   \\
\hline
\end{tabular}
  \caption{\textbf{Generalization of deep networks in classification (accuracy $\pm$ std.).} When the models are initialized with higher bandwidths (pre-trained on random labels), the test accuracy drops. This pattern is consistent across various architectures and datasets. We do not use data augmentation in these experiments and each model is run for five times in each setting.}
  \label{tab:deep_networks}
\end{table}

\section{A case against the flat minima conjecture}

The flat minima conjecture speculates that the generalisation capacity of a network is related to loss landscape geometry. Simply put, the flat minima conjecture postulates that -- for some definition of the flatness of the loss landscape -- flat minima lead to better generalization (but not vice-versa \citep{sharpminima}). Next, we show that while this conjecture is consistent with the evidence for ReLU networks, it is falsified by Gaussian networks.

\begin{figure}[h]
\includegraphics[width=1.0\columnwidth]{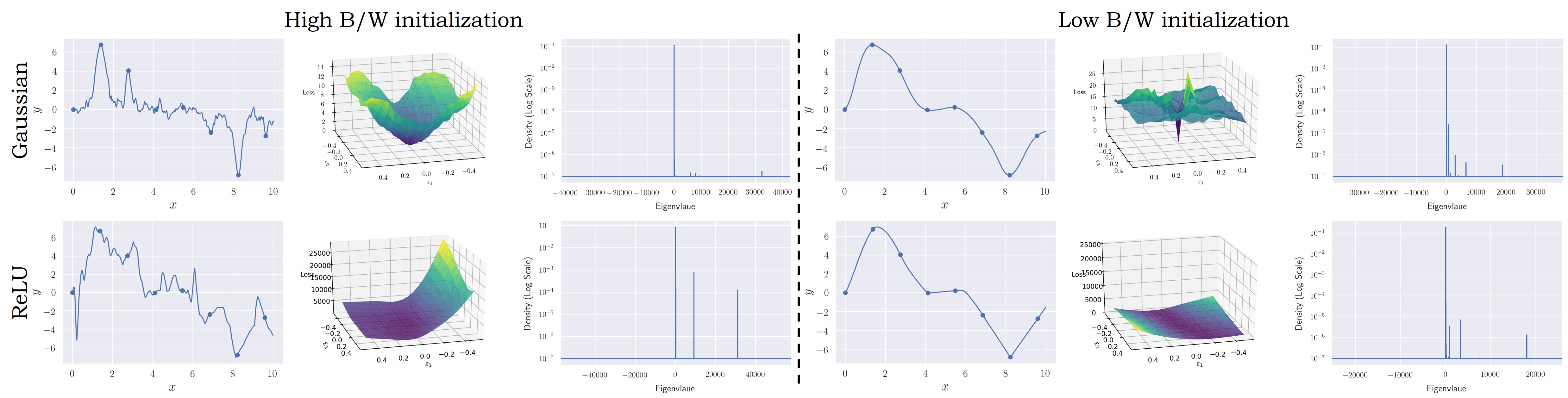}
\vspace{-2em}
\caption{\small \textbf{Flat minima conjecture does not always hold.} The left block and the right block correspond to high bandwidth and low bandwidth initializations, respectively. In each block, from the left column, the interpolations, loss landscapes, and the eigenvalue distribution of the loss-Hessian are illustrated. The loss landscapes are plotted along with the directions of the two largest eigenvalues. As depicted, while our results for the ReLU network are consistent with the conjecture, the Gaussian network behaves in the opposite manner. For more detailed quantitative results, see Table \ref{tab:uneven}.}
\label{fig:flat}
\end{figure}

\paragraph{Experiment 5} We sample four random variables $w_1, w_2 \sim U(0.01, 1), a_1,a_2 \sim U(1,10)$ and define $20$ signals using the sampled variables as $a_1\mathrm{sin}(2\pi w_1x) + a_2\mathrm{sin}(2\pi w_2x)$. Then, we sample $8$ equidistant samples between $0$ and $10$, and use them as the training points to train models. We use Gaussian and ReLU networks for this experiment in two settings. In the first setting, we initialize the ReLU network using Xavier initialization and the Gaussian networks with $\mathcal{N}(0,0.03)$. In this setting, both the networks are able to interpolate the points smoothly. In the other setting, we initialize the ReLU network by pre-training it on $\mathrm{sin}(6\pi x)$ and the Gaussian network with Xavier initialization. In this scenario, both networks demonstrate non-smooth interpolations due to initial high bandwidth. At convergence, we compute the hessian of the loss with respect to the parameters and then compute the eigenvalues and the trace of the hessian. The results are shown in Fig.~\ref{fig:flat} and Table~\ref{tab:uneven}. As evident, the behavior of the Gaussian network is not consistent with the flat minima conjecture. Thus, we advocate that the spectral bias provides a more reliable framework for investigating the generalization of neural networks.

\begin{table}
\scriptsize
\centering
       \begin{tabular}{||c|c|c|c|c|c||}
\hline
Model & Initialization & Hessian trace & $\mathbb{E}[\epsilon]$ & Spectral norm\\
\hline
\hline
ReLU & High B/W & $134213.36$ & $0.95$ & $257875.23$\\
ReLU & Low B/W & $31110.73$ & $0.04$ & $49781.58$\\

Gaussian & High B/W & $40478.82$  & $0.21$ & $12596.89$\\

Gaussian & Low B/W & $59447.46$ & $0.32$ & $26519.66$\\
\hline
\end{tabular}
\caption{\small The trace,  expected eigenvalue ($\mathbb{E}[\epsilon)$], and the spectral norm of the loss-Hessian are shown (averaged over $20$ signals). Higher values indicate a sharper minimum. As illustrated, while the ReLU network obeys the flat minima conjecture, the Gaussian network behaves oppositely.}
\label{tab:uneven}
\end{table}

\section{Conclusion} 
\vspace{-1em}
In this paper, we focus on the effect of initialization on the implicit generalization of neural networks. In particular, we reveal that the Fourier spectrum of the network at initialization has a significant impact on the generalization gap. Moreover, we offer evidence against the flat minima conjecture and show that the correlation between the flatness of the minima and the generalization can be architecture-dependent. We empirically validate the generality of our insights across diverse, practical settings.

\bibliographystyle{plainnat}
\bibliography{main}

\section*{Checklist}



\begin{enumerate}

\item For all authors...
\begin{enumerate}
  \item Do the main claims made in the abstract and introduction accurately reflect the paper's contributions and scope?
    \answerYes{}
  \item Did you describe the limitations of your work?
    \answerYes{}
  \item Did you discuss any potential negative societal impacts of your work?
    \answerNA{}
  \item Have you read the ethics review guidelines and ensured that your paper conforms to them?
    \answerYes{}
\end{enumerate}

\item If you are including theoretical results...
\begin{enumerate}
  \item Did you state the full set of assumptions of all theoretical results?
   \answerYes{}
        \item Did you include complete proofs of all theoretical results?
     \answerYes{}
\end{enumerate}

\item If you ran experiments...
\begin{enumerate}
  \item Did you include the code, data, and instructions needed to reproduce the main experimental results (either in the supplemental material or as a URL)?
     \answerNo{}
  \item Did you specify all the training details (e.g., data splits, hyperparameters, how they were chosen)?
    \answerYes{}
        \item Did you report error bars (e.g., with respect to the random seed after running experiments multiple times)?
    \answerNA{}
        \item Did you include the total amount of compute and the type of resources used (e.g., type of GPUs, internal cluster, or cloud provider)?
    \answerNo{}
\end{enumerate}

\item If you are using existing assets (e.g., code, data, models) or curating/releasing new assets...
\begin{enumerate}
  \item If your work uses existing assets, did you cite the creators?
    \answerYes{}
  \item Did you mention the license of the assets?
    \answerNA{}
  \item Did you include any new assets either in the supplemental material or as a URL?
    \answerNA{}
  \item Did you discuss whether and how consent was obtained from people whose data you're using/curating?
    \answerNA{}
  \item Did you discuss whether the data you are using/curating contains personally identifiable information or offensive content?
    \answerNA{}
\end{enumerate}

\item If you used crowdsourcing or conducted research with human subjects...
\begin{enumerate}
  \item Did you include the full text of instructions given to participants and screenshots, if applicable?
    \answerNA{}
  \item Did you describe any potential participant risks, with links to Institutional Review Board (IRB) approvals, if applicable?
    \answerNA{}
  \item Did you include the estimated hourly wage paid to participants and the total amount spent on participant compensation?
    \answerNA{}
\end{enumerate}

\end{enumerate}
\newpage


\appendix

\section{Appendix}

\subsection{Proof of Theorem 5.1}

\begin{proof}
The evolution equation for $\mathcal{F}[g_{\theta(t)}]$ follows easily from the Liebniz integral rule:
\[
\frac{d}{dt}\mathcal{F}[g_{\theta(t)}] = \mathcal{F}\bigg[\frac{d}{dt}g_{\theta(t)}\bigg].
\]
Now, $\mathcal{K}$ is nothing other than the extension of the tangent kernel associated to $f_{\theta}$ by zero outside of the compact neighbourhood $K$ of the data manifold, i.e.
\[
\mathcal{K}(\theta,x,x') = \begin{cases} D_{\theta}f_{\theta}(x)\,D_{\theta}f_{\theta}(x')^{T},&\text{ if $x,x'\in K$}\\ 0,&\text{ otherwise.}\end{cases}
\]
By hypothesis on $f_{\theta}$, one has that $\frac{d}{dt}g_{\theta(t)}$ is an $L^1$ function, so that by the Riemann-Lebesgue lemma its Fourier transform vanishes at infinity as stated.
\end{proof}

\subsection{Initializing deep networks with higher bandwidths}

Initializing deep classification networks -- that consume high dimensional inputs such as images -- such that they have higher bandwidths is not straightforward. Therefore, we explore alternative ways to initialize networks with higher bandwidths in low-dimensional settings, and extrapolate the learned insights to higher dimensions.

\begin{figure}[!htp]
    \centering
    \includegraphics[width=1.\columnwidth]{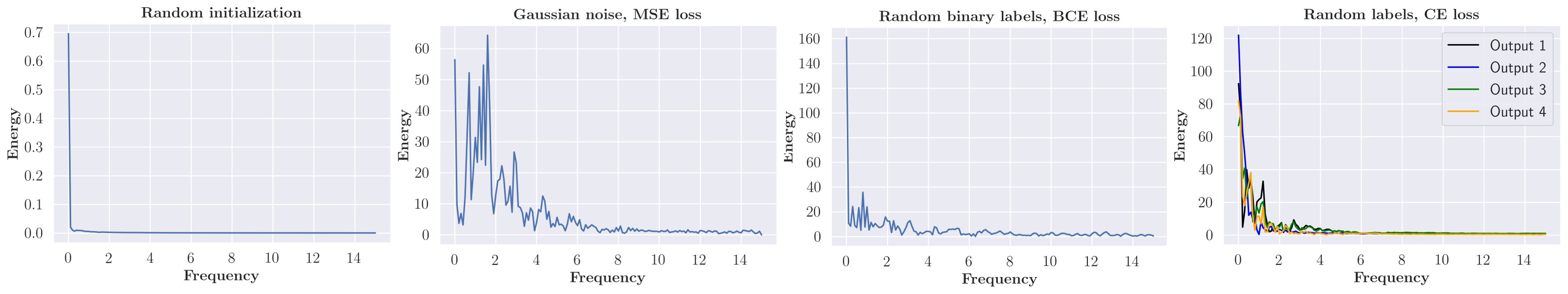}
    \caption{\small We visualize the spectra of networks after training them with different loss functions and label sampling schemes (the rightmost three plots). All shown methods are able to obtain higher bandwidths than random initialization (leftmost plot). Note that the scale in the $y-$axis is different for each plot. However, in practice, deep classification networks take an infeasible amount of time to converge with MSE loss. Hence, we chose random labels with cross-entropy loss to initialize the deep classification networks with higher bandwidths.}
    \label{fig:intialization1}
\end{figure}

For all the experiments, we consider a fully connected 4-layer ReLU network with 1-dimensional inputs. First, we sample a set of values from white Gaussian noise, and train the network with these target values using MSE loss. In the second experiment, we threshold the sampled values to obtain a set of binary labels, and then train the network with binary cross-entropy loss. For the third experiment, we use a network with four outputs. Then, we separate the sampled values into four bins, and obtain four labels. Then, we train the network with cross-entropy loss. We compute the Fourier spectra of each of the trained networks after convergence. The results are shown in Fig.\ref{fig:intialization1}. 

As depicted, we can use mean squared error (MSE) or cross-entropy (CE) loss along with random labels to initialize the networks with higher bandwidth. However, we observed that, in practice, deep networks take an infeasible amount of time to converge with the MSE loss. Therefore, we use cross-entropy loss with random labels to initialize the networks in image classification settings.

In order to verify that training with random labels indeed induces higher bandwidths on deep classification networks, we visualize the histograms of their first order gradients of the averaged outputs w.r.t. the inputs. It is straightforward to show that (similar to second-order gradients) higher first-order gradients lead to higher bandwidth. For simplicity, consider a  function $f:\mathbb{R} \to \mathbb{R}$. Then,

\[
   f(x) =  \int_{\infty}^{\infty} \hat{f}(k)e^{2\pi ikx}dk
\]
It follows that,

\begin{align}
    |\frac{df(x)}{dx}| & =      |2\pi i \int_{\infty}^{\infty} k\hat{f}(k)e^{2\pi ikx}dk|\\
    & \leq |2 \pi| \int_{\infty}^{\infty} |k\hat{f}(k)|dk.
\end{align}

Therefore,

\begin{equation}
    \max_{x \in \epsilon} |\frac{df(x)}{dx}| \leq |2 \pi| \int_{\infty}^{\infty} |k\hat{f}(k)|dk.
\end{equation}

\begin{figure}[!htp]
    \centering
    \includegraphics[width=1.\columnwidth]{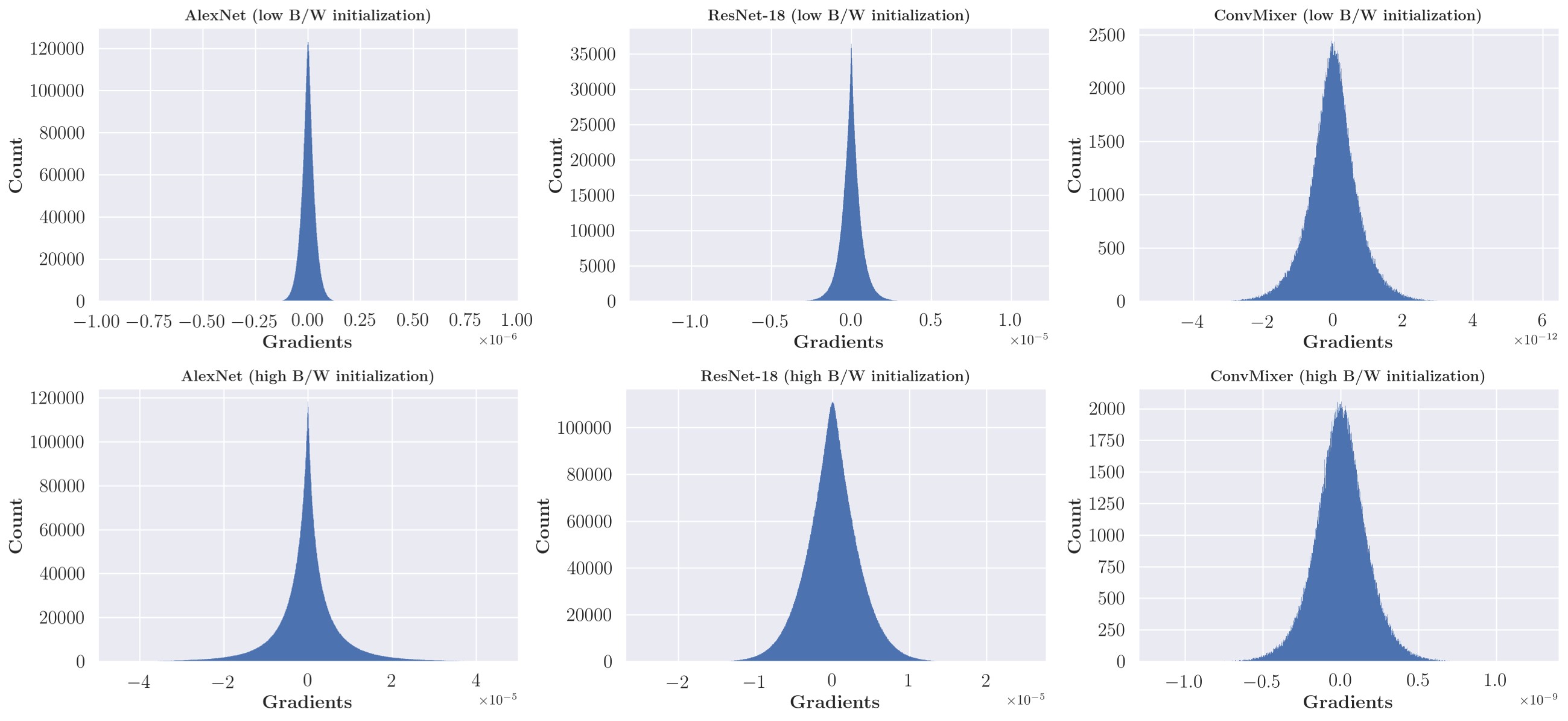}
    \caption{\small The histograms of the first-order gradients of the outputs with respect to the inputs (a batch of training images) are shown. Low and high bandwidth initializations correspond to Xavier initialization and pre-training with random labels, respectively. Not that the $x-$axis scales are different in each plot. As depicted, training with random labels leads to higher gradients, validating that it indeed leads to higher bandwidths.}
    \label{fig:gradients}
\end{figure}

This conclusion can be directly extrapolated to higher-dimensional inputs, where the Fourier transform is also high dimensional. Hence, we feed a batch of images to the networks, and calculate the gradients of the averaged output layer with respect to the input image pixels. Then, we plot the histograms of the gradients 
(Fig.~\ref{fig:gradients}). As illustrated, training with random labels induces higher gradients, and thus, higher bandwidth. Table.~\ref{tab:deep_networks_imagenet} compares generalization of deep networks on ImageNet.

\begin{table}[!htp]
\scriptsize
\centering
       \begin{tabular}{||c|c|c|c|c||}
\hline

\multicolumn{5}{||c||}{ImageNet} \\

\hline
\multirow{ 2}{*}{Model} & 
\multicolumn{2}{|c|}{Random initialization}  &  \multicolumn{2}{|c||}{High B/W initialization} \\
\cline{2-5}
 & 
 Train accuracy & Test accuracy & Train accuracy &  Test accuracy \\
\hline
VGG16              & $100\%$ & $68.19\%$ & $100\%$ & $55.48$\%\\
ResNet-18  & $100\%$ & $66.93\%$                   & $100\%$ & $49.17\%$\\
ConvMixer        & $100\%$ & $74.19\%$                   & $100\%$ & $45.68\%$\\
\hline
\end{tabular}
  \caption{\textbf{Generalization of deep networks in classification over ImageNet.} When the models are initialized with higher bandwidths (pre-trained on random labels), the test accuracy drops. We do not use data augmentation in these experiments. We only use three models for this experiment due to the extensive resource usage when training on random labels over ImageNet.}
  \label{tab:deep_networks_imagenet}
\end{table}

\subsection{Convergence-decay rates of frequencies matter for generalization}

Earlier, we showed that although all neural networks admit spectral bias, the convergence-decay rates of frequencies change across network types and initialization schemes. Below, we show that these decay rates play an essential role in generalization.

\begin{figure}[!htp]
    \centering
    \includegraphics[width=1.\columnwidth]{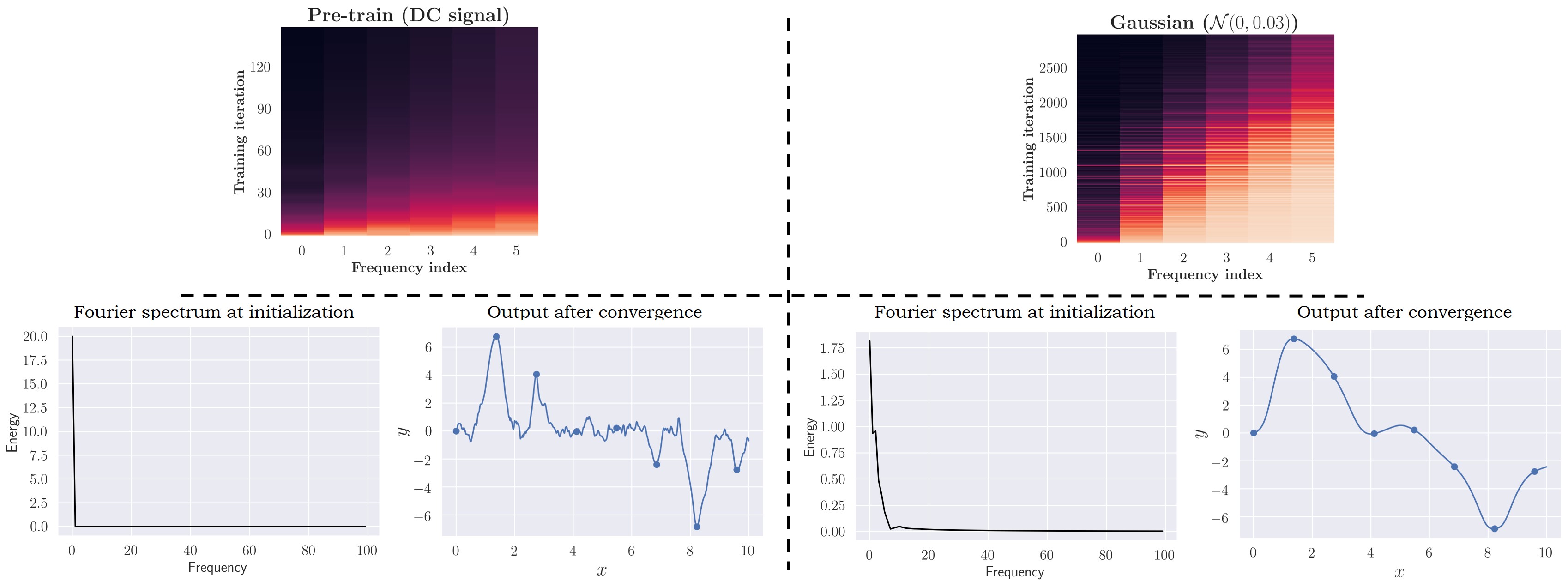}
    \caption{\small \textbf{The effect of convergence-decay rate of frequencies on generalization.} \textit{Left block}: We pre-train a Gaussian network on a DC signal to obtain low initial bandwidth. Nevertheless, the network still converges to a non-smooth solution. \textit{Right block}: The Gaussian network is initialized using a random  Gaussian distribution ($\mathcal{N}(0,0.03)$). This method also leads to lower bandwidth. However, in this scenario, the network is able to converge to a smooth solution. At the top, the convergence of frequency components -- starting from the corresponding initialization -- is shown when training on a signal $g(x) = \sum_{n=1}^{6} \mathrm{sin}(10\pi nx )$. Note that a lower convergence decay rate leads to bad generalization.}
    \label{fig:architecture}
\end{figure}

We use a Gaussian network for this experiment. We initialize two instances of the network by 1) using a weight distribution $\mathcal{N}(0,0.03)$, and 2) pre-training the network on a DC signal. In both instances, the network has low bandwidth. Then, we train the network on sparse training data sampled from $3\mathrm{sin}(0.4\pi x) + 5\mathrm{sin}(0.2\pi x)$. The results are shown in Fig.~\ref{fig:architecture}. Observe that although both networks start from low bandwidth, they exhibit different generalization properties. This is because, having a lower convergence-decay hinders smooth interpolations even in cases where the networks have low initial bandwidth. This is expected, since then, the optimization will begin to affect the higher frequencies before the lower frequencies are converged. 

\begin{figure}[!htp]
    \centering
    \includegraphics[width=1.\columnwidth]{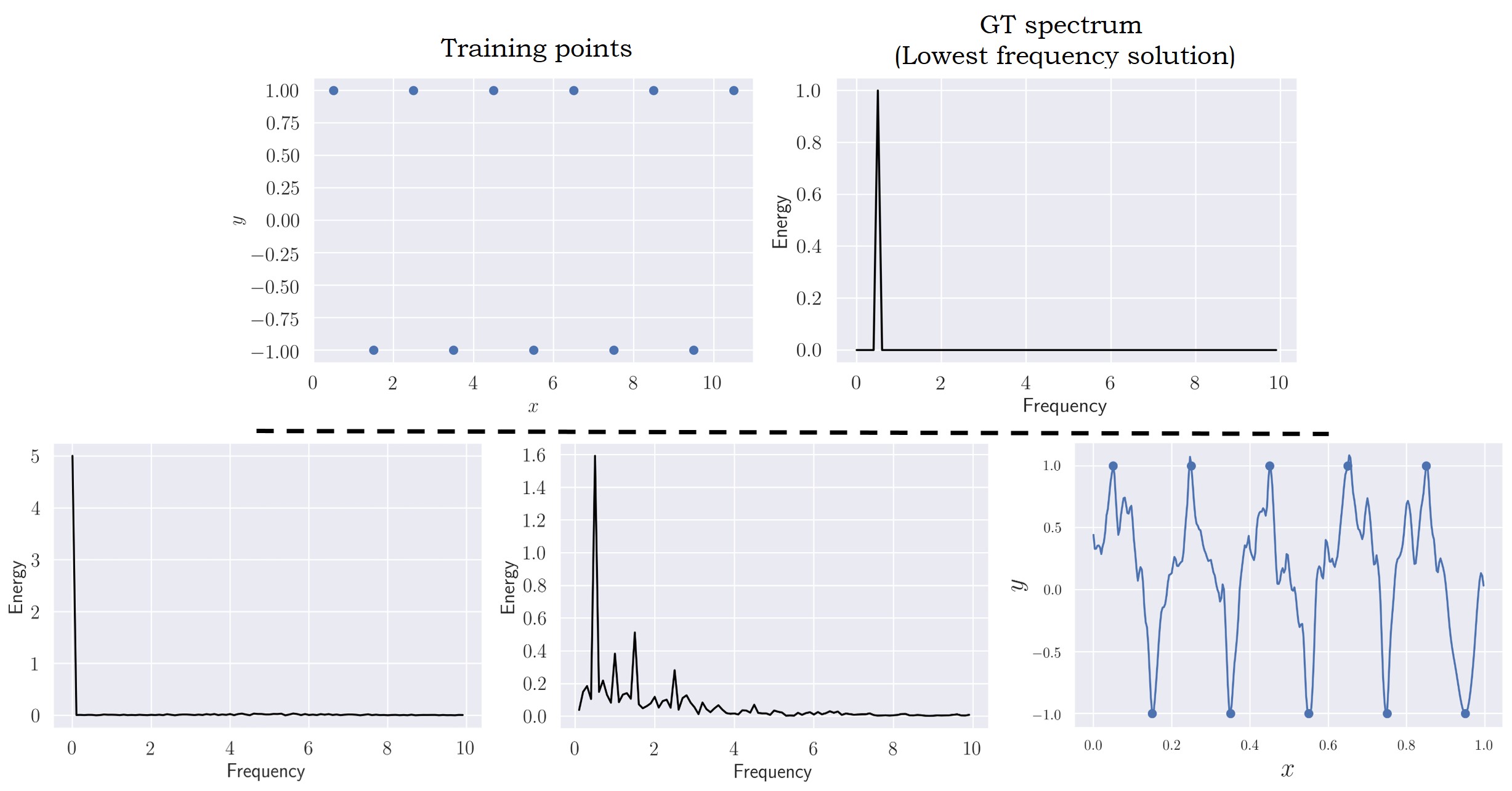}
    \caption{\small The top block shows sparsely sampled training points from $\mathrm{sin}(\pi x)$ and the corresponding lowest frequency solution that fits the training data. The bottom block shows the spectra of a Gaussian network initialized by pre-training on a DC signal. Even though the network adds a spike at the lowest frequency solution, higher frequencies are also added to the spectrum due to the low convergence-decay rate. This results in a non-smooth interpolation.}
    \label{fig:gaussianlowest}
\end{figure}

To further verify this, we conduct another experiment; see Fig.~\ref{fig:gaussianlowest}.

\subsection{Analyzing the loss landscapes}

\begin{figure}[!htp]
    \centering
    \includegraphics[width=1.\columnwidth]{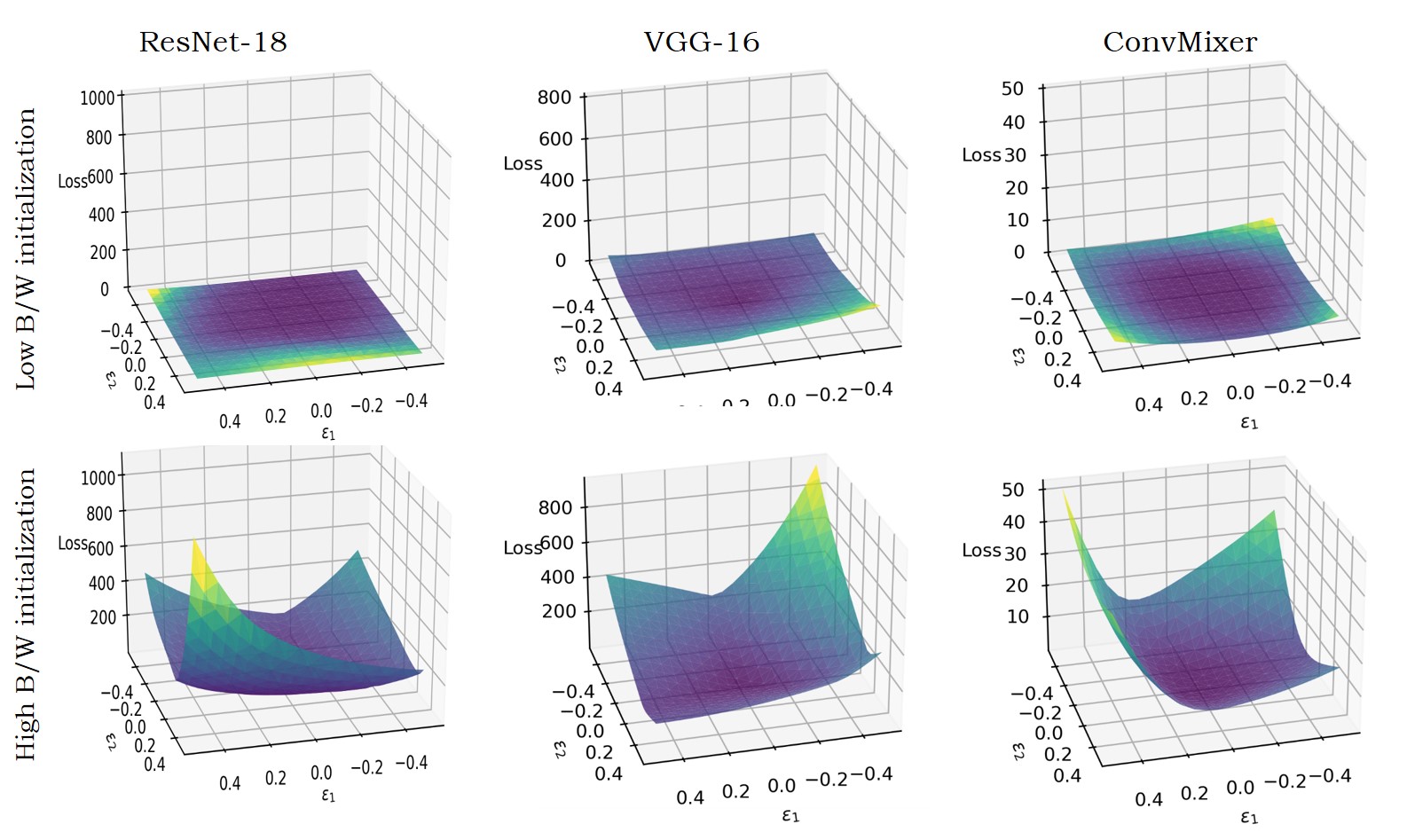}
    \caption{\small \textbf{Loss landscapes of deep networks trained on CIFAR10.} The proposed high B/W initialization scheme provides an ideal platform to compare the flatness of minima with different generalization properties. Note that ReLU networks exhibit behaviour consistent with the flat minima conjecture.}
    \label{fig:loss_vis}
\end{figure}

The flat minima conjecture has been studied since the early work of \citet{hochreiter1994simplifying} and \citet{hochreiter1997flat}. More recently, empirical works showed that the generalization of deep networks is related to the flatness of the minima it is converged to during training \citep{chaudhari2019entropy, keskar2016large}. In order to measure the flatness of loss landscapes, different metrics have been proposed \citep{tsuzuku2020normalized, rangamani2019scale, hochreiter1994simplifying, hochreiter1997flat}. In particular, \citet{chaudhari2019entropy} and \citet{keskar2016large} showed that minima with low Hessian spectral norm generalize better.  In this paper also, we use Hessian-related metrics to measure the flatness. Since the spectral norm alone is not ideal for analyzing the loss landscape of models with a large number of parameters, we also compute the trace and the expected eigenvalue of the Hessian. For computing the Hessian, we use the library provided by \citet{yao2020pyhessian}. Fig~\ref{fig:loss_vis} and Table \ref{tab:minmia} depict a comparison of loss landscapes in several deep models. Note that our proposed high B/W initialization scheme provides an ideal platform to compare the loss landscapes with different generalization properties.

\begin{table}[!htp]
    \centering
 \begin{tabular}{||c|c|c||}

\hline
Model & Hessian-trace &  Spectral norm  \\
\hline
ResNet-18 (low B/W) & $13560.76$ & $2805.47$\\
ResNet-18 (high B/W) & $28614.19$ & $4121.36$\\
VGG-16 (low B/W) &$10102.51$ & $1112.07$   \\
VGG-16 (high B/W) &$14483.90$ & $3214.57$   \\
ConvMixer (low B/W) &  $0.3242$ & $0.028$ \\
ConvMixer (high B/W) &  $3.49$ & $0.445$ \\
\hline
\end{tabular}
 \caption{\small Quantitative comparison of the flatness of minima in deep networks. Note that Note that ReLU networks exhibit behaviour consistent with the flat minima conjecture.}
\label{tab:minmia}
\end{table}

\end{document}